\documentclass[conference,flushend]{iaria} 
\usepackage{amsmath,amssymb,amsfonts}
\usepackage{algorithm}
\usepackage{graphicx}
\usepackage{textcomp}
\usepackage{xcolor}
\usepackage{hyperref}
\usepackage{textcase}
\usepackage[tablename=TABLE]{caption}
\DeclareCaptionTextFormat{up}{\MakeTextUppercase{#1}}
\DeclareCaptionLabelFormat{up}
{\MakeTextUppercase{#1} #2}
\usepackage{subcaption}
\usepackage[per-mode=fraction]{siunitx}
\graphicspath{ {./figs/} }
\def\BibTeX{{\rm B\kern-.05em{\sc i\kern-.025em b}\kern-.08em
    T\kern-.1667em\lower.7ex\hbox{E}\kern-.125emX}}

\usepackage[nolist]{acronym}
\usepackage{orcidlink}
\usepackage[english]{babel}
\usepackage{randtext}
\newcommand{\email}[1]{
  \texttt{\randomize{#1}}
}

\usepackage{pgfplotstable}
\usepackage{siunitx}
\usepackage{listings}
\usepackage{pgfplots}
\usepackage{tikz}
\usepackage{float}
\usepackage{algpseudocode}
\usepgfplotslibrary{groupplots}

\usepackage{listings}
\usepackage{xcolor}

\title{A PyTorch-Compatible Spike Encoding Framework for Energy-Efficient Neuromorphic Applications}
\author{
    \IEEEauthorblockN{
        Alexandru Vasilache$^{1,2}$,
        Jona Scholz$^{1}$,
        Vincent Schilling$^{2}$,
        Sven Nitzsche$^{1,2}$,\\
        Florian Kaelber$^{3}$,
        Johannes Korsch$^{3}$,
        Juergen Becker$^{2}$
    }
    
    \IEEEauthorblockA{
        $^{1}$ \textit{FZI Research Center for Information Technology, Karlsruhe, Germany}\\ 
        $^{2}$ \textit{Karlsruhe Institute of Technology, Karlsruhe, Germany} \\
        $^{3}$ \textit{NXP Semiconductors Germany GmbH, Munich, Germany} \\
        \email{email: \{vasilache, jona.scholz, nitzsche\}@fzi.de} \\
        \email{utmlb@student.kit.edu} \email{Juergen.Becker@kit.edu} \\
        \email{\{florian.kaelber, johannes.korsch\}@nxp.com} \\
    }
}
\begin{document}


\maketitle
\begin{acronym}[Longest Abrev] 
\acro{ANN}{Artificial Neuronal Network}
\acro{AE}{Auto-Encoder}
\acro{CNN}{Convolutional Neural Network}
\acro{DNN}{Deep Neural Network}
\acro{LIF}[LIF]{Leaky Integrate-and-Fire}
\acro{IF}[IF]{Integrate-and-Fire}
\acro{LI}[LI]{Leaky Integrate}
\acro{ALIF}{Adaptive Leaky Integrate-and-Fire}
\acro{adex}[AdEx]{adaptive exponential integrate-and-fire}
\acro{LSTM}{Long Short-Term Memory}
\acro{RNN}{Reccurent Neural Network}
\acro{LSNN}{Long Short-Term Spiking Neural Network}
\acro{NN}{Neural Network}
\acro{SNN}{Spiking Neural Network}
\acro{ML}{Machine Learning}
\acro{DL}{Deep Learning}
\acro{AI}{artificial intelligence}
\acro{DFT}{Discrete Fourier Transform}
\acro{FFT}{Fast Fourier Transform}
\acro{STFT}{Short-Time Fourier transform}
\acro{BPTT}{Backpropagation Through Time}
\acro{BP}{Backpropagation}
\acro{LMD}{Local Mean Decomposition}
\acro{GRF}{Gaussian Receptive Field}
\acro{TTFS}{Time-to-First-Spike}
\acro{PCA}{Principal component analysis}
\acro{FPGA}[FPGA]{Field Programmable Gate Array}
\acro{SF}[SF]{Step-Forward}
\acro{SW}[SW]{Sliding Window}
\acro{MW}[MW]{Moving Window}
\acro{BSA}[BSA]{Ben's Spiker Algorithm}
\acro{PWM}[PWM]{Pulse Width Modulation}
\acro{PWMB}[PWMB]{Pulse Width Modulated-Based}
\acro{RL}[RL]{Reinforcement Learning}
\acro{FIR}[FIR]{Finite Impulse Response}
\acro{MSE}[MSE]{Mean Squared Error}
\acro{GRF}[GRF]{Gaussian Receptive Field}
\end{acronym}

\begin{abstract} 
Spiking Neural Networks (SNNs) offer promising energy efficiency advantages, particularly when processing sparse spike trains. However, their incompatibility with traditional datasets, which consist of batches of input vectors rather than spike trains, necessitates the development of efficient encoding methods. This paper introduces a novel, open-source PyTorch-compatible Python framework for spike encoding, designed for neuromorphic applications in machine learning and reinforcement learning. The framework supports a range of encoding algorithms, including Leaky Integrate-and-Fire (LIF), Step Forward (SF), Pulse Width Modulation (PWM), and Ben’s Spiker Algorithm (BSA), as well as specialized encoding strategies covering population coding and reinforcement learning scenarios. Furthermore, we investigate the performance trade-offs of each method on embedded hardware using C/C++ implementations, considering energy consumption, computation time, spike sparsity, and reconstruction accuracy. Our findings indicate that SF typically achieves the lowest reconstruction error and offers the highest energy efficiency and fastest encoding speed, achieving the second-best spike sparsity. At the same time, other methods demonstrate particular strengths depending on the signal characteristics. This framework and the accompanying empirical analysis provide valuable resources for selecting optimal encoding strategies for energy-efficient SNN applications.
\end{abstract}


\section{Introduction}

An increasing number of tasks rely on \ac{ML} to enhance accuracy, streamline development processes, or facilitate automation in the first instance. \acp{NN} are often employed for this purpose due to their flexibility and capacity to solve even the most complex tasks. As neural networks become more widespread, they are also being employed with increasing frequency in embedded systems. However, their deployment in such embedded contexts is frequently constrained by the requisite computing power and the associated high energy demands. 
Given these considerations, SNNs may offer a promising avenue for integrating high-performance machine learning into embedded systems, provided that suitable neuromorphic hardware is available. 

In order to utilize SNNs, it is necessary to have access to adequate spiking data, also referred to as event-based data. Moreover, the event-based data must be sparse to leverage the advantages of spiking neural networks fully. Ideally, an event-based sensor would be employed to provide this data. However, there is a notable scarcity of such sensors on the market. Currently, the only commercially available event-based sensors target the vision modality through event-based cameras \cite{aerne_inivation_nodate}\cite{noauthor_prophesee_nodate}.

As an alternative approach, non-spiking data can be converted using spike encoding algorithms. Such algorithms are designed to transform real-valued data into spike trains, which can then be utilized as input for an SNN. 
The existing literature classifies spike encoding algorithms into two principal categories: rate coding and temporal coding \cite{auge2021survey}. In rate coding, the signal amplitudes are directly mapped to spike frequencies, resulting in high spiking activity. In contrast, temporal coding results in the firing of spikes only when specific events occur, thereby making the timing of spikes crucial and reducing the overall number of spikes. This sparsity renders temporal coding more power-efficient and suitable for low-power applications. This paper focuses on evaluating four temporal encoding methods: \Ac{LIF}, \Ac{SF}, \Ac{PWM}, and \Ac{BSA}. These methods were chosen as they are frequently discussed and compared in spike encoding literature \cite{wang2023comparison}\cite{chen2022temporal}\cite{petro2019selection}\cite{yarga2022efficient}, serving as representative techniques for this study.

Even though many spike encoding methods are publicly available, no open-source framework or library currently groups various such algorithms and allows for straightforward integration into popular machine-learning workflows. Accordingly, this work seeks to provide a framework that offers out-of-the-box PyTorch support and automatic parameter optimization of the encoding algorithms for a given dataset. Additionally, we evaluate the performance of various spike encoding algorithms for specific signal types and implement them on an embedded platform to assess runtime and power demand. The library is publicly accessible at its GitHub Repository \cite{vasilache_alex-vasilachespike-encoding_2025}. This work aims to reduce the overhead associated with integrating SNNs into real-world applications. 

The remainder of this paper is organized as follows: Section~\ref{sec:related-work} compares our work with existing state-of-the-art approaches. Section~\ref{sec:methods} then summarizes the investigated methods and details the experimental setup, followed by Section~\ref{sec:results} which displays the results regarding reconstruction error, spike sparsity, runtime, and power consumption. Subsequently, Section~\ref{sec:discussion} discusses these results focusing on energy efficiency and accuracy for different signal types. Finally, Section~\ref{sec:repository} outlines the structure of the provided spike encoding library.


\section{Related Work}
    \label{sec:related-work}

\subsection{Evaluation of Spike Encoding Algorithms}
Wang et al. \cite{wang2023comparison} implemented four spike encoding algorithms on a \ac{FPGA} and compared them by their speed, power consumption, accuracy, and robustness to noise. They evaluated \ac{SW} encoding \cite{webb2011spiking}, \ac{BSA} \cite{schrauwen2003bsa}, \ac{SF} encoding \cite{kasabov2016evolving} and the \ac{PWM} algorithm \cite{arriandiaga2019pulsewidth}. Furthermore, they verified their evaluation results on two real-world applications: tactile signal encoding reconstruction and a robotic arm control task. They found that overall \ac{BSA} had the highest power consumption and was therefore not recommended for most applications, with the notable exception of encoding square signals, where its energy efficiency was high. \Ac{PWM} performed well in most signal reconstruction tests, and its simple implementation made it favorable for real-world use. However, its accuracy may be decreased if an unsuitable curve-fitting algorithm is chosen. Also, if non-linear operations are used, the power consumption may be elevated. This added variability is reduced in \Ac{SF} encoding, which only has one adjustable parameter. While \ac{SF} encoding also performed well in most signal reconstruction tests, its accuracy could degrade when high changes in the signal amplitude occurred. Finally, \ac{SW} encoding had no notable advantages over the other algorithms. 

Chen et al. \cite{chen2022temporal} compared (among others) \ac{SF} to \ac{PWM} and provided further evidence that both algorithms achieve high signal reconstruction accuracy. They observed that \ac{SF} is more accurate at lower thresholds, but the increased spike count may also increase power consumption. 

In \cite{petro2019selection}, the authors suggest a workflow for selecting, optimizing, and validating spike encoding methods, which they evaluate on \ac{SF}, \ac{BSA}, and others. Their experiments provide further evidence for \acp{SF} versatility and robustness, consistent with similar studies' findings. However, their evaluation of \ac{BSA} contradicts those of \cite{wang2023comparison} regarding step signals since they observed high signal reconstruction errors for this signal type. An implementation choice may partly explain this since they use a multiplicative threshold rather than a subtractive one. 

Yarga et al. \cite{yarga2022efficient} applied \ac{BSA}, \ac{LIF} encoding, and three additional methods to encode voice recordings. Their results are not fully applicable to our comparison, as they focused on the accuracy of a classification task rather than evaluating reconstruction error. Initially, features were extracted into a spectrogram or cochleagram and then converted to spike trains using the evaluated encoding methods. These spike trains were subsequently processed by a \ac{CNN} for classification. 
Their evaluation metrics included spike density and classification accuracy, which are indirectly related to the energy efficiency of the encoding and its impact on information loss or gain. However, these metrics may not fully represent the performance of the encoding methods. Most methods allowed spike density to be adjusted by modifying their parameters, such as the membrane threshold in \ac{LIF} encoding. On spectrogram features, their findings showed that \ac{LIF} encoding produced the highest accuracy across most spike densities among the evaluated methods. Notably, when using spectrogram features, \ac{BSA} also performed well at very low spike densities. However, on cochleagram features, \ac{BSA} performed poorly, and \ac{LIF} was outperformed by other methods at low spike densities. At higher spike densities, \ac{LIF} once again achieved the highest overall classification accuracy. In summary, their findings suggest that \ac{LIF} encoding can achieve both high classification accuracy and low spike densities, contributing to better energy efficiency. In contrast, \ac{BSA} is more restricted to specific scenarios where it performs well.


\subsection{Spike Encoding Repositories}
The currently available spike encoding repositories tend to prioritize applications outside the domain of machine learning. 
SpikeCoding \cite{dupeyroux2022toolbox} emphasizes the real-time control of robots and has been developed to integrate with ROS rather than focusing on machine learning workflows. 
Similarly, Spikes \cite{gollahalli_akshaybabloospikes_2024} offers foundational tools for spike generation but is not designed for integration with neural network models. 
In contrast, the proposed framework addresses this limitation by providing a PyTorch-compatible, open-source library focused on machine learning and neural network applications, including support for reinforcement learning environments, facilitating broader use in machine learning research and practice.

\section{Methods}
    \label{sec:methods}
This section presents the investigated encoding methods and their evaluation on an embedded hardware platform and outlines the experimental setup. This includes an explanation of the underlying hardware processes, the types of signals used, and the parameter optimization of the encoding algorithms.

\subsection{Encoding Algorithms}
\Ac{SF} encoding is a straightforward and efficient method for signal processing. A spike is generated whenever the signal surpasses a defined baseline, which is subsequently incremented by a constant value. Conversely, if the signal did not exceed the baseline, the baseline is lowered by a constant value. A pseudocode implementation is shown in Figure \ref{fig:SF_encoding} (based on \cite{SF_SOURCE_kasabov2016evolving}). Encoding spikes in this way is similar to the \ac{LIF} encoding method, although the latter is inspired by a spiking neuron model of the same name. 


\begin{figure}[htbp] 
    \centering 
    \begin{minipage}{1\linewidth} 
        \footnotesize 
        \begin{algorithmic}[1] 
            \State \textbf{Input:} $signal$, $threshold$
            \State \textbf{Output:} $spike\_train$
            \State $base \gets 0$, $up\_spikes \gets 0$, $down\_spikes \gets 0$
            \For{$t = 1$ to $time\_steps$}
                \If{$signal(t) > base + threshold$}
                    \State $up\_spikes(t) \gets 1$
                    \State $base \gets base + threshold$
                \ElsIf{$signal(t) < base - threshold$}
                    \State $down\_spikes(t) \gets -1$
                    \State $base \gets base - threshold$
                \Else
                    \State $up\_spikes(t) \gets 0$
                    \State $down\_spikes(t) \gets 0$
                \EndIf
            \EndFor
            \State $spikes \gets up\_spikes + down\_spikes$
        \end{algorithmic}
    \end{minipage} 
    \caption{SF Encoding Method} 
    \label{fig:SF_encoding} 
\end{figure}

In \ac{LIF} encoding, the signal serves as an input current that increases the membrane potential. When the potential exceeds a predefined threshold, a spike is emitted. The membrane potential decays proportionately to a decay variable at each time step. It is important to note that the original signal must be normalized, as neither the decay rate nor the threshold adapts to the varying range of possible signal values. A pseudocode implementation of the \ac{LIF} encoding method is provided in Figure \ref{fig:LIF_encoding} (based on \cite{yarga2022efficient}), where $\text{min\_max\_normalize}$ refers to a rescaling of the range of features to $[0, 1]$.


\begin{figure}[htbp] 
    \centering 
    \begin{minipage}{1\linewidth} 
        \footnotesize 
        \begin{algorithmic}[1] 
            \State \textbf{Input:} $signal$, $threshold$, $membrane\_constant$
            \State \textbf{Output:} $spike\_train$
            \State $signal \gets \text{min\_max\_normalize}(signal)$
            \State $signal \gets signal \times 2 - 1$
            \State $voltage \gets 0$, $up\_spikes \gets 0$, $down\_spikes \gets 0$
            \For{$t = 1$ to $time\_steps$}
                \State $voltage \gets voltage + signal(t)$
                \If{$voltage > threshold$}
                    \State $up\_spikes(t) \gets 1$
                    \State $voltage \gets 0$
                \ElsIf{$voltage < -threshold$}
                    \State $down\_spikes(t) \gets -1$
                    \State $voltage \gets 0$
                \Else
                    \State $up\_spikes(t) \gets 0$
                    \State $down\_spikes(t) \gets 0$
                \EndIf
                \State $voltage \gets voltage \times membrane\_constant$
            \EndFor
            \State $spikes \gets up\_spikes + down\_spikes$
        \end{algorithmic}
    \end{minipage} 
    \caption{LIF Encoding Method} 
    \label{fig:LIF_encoding} 
\end{figure}

Similar to \ac{LIF} encoding, \ac{BSA} requires data normalization but employs a fundamentally different approach. Instead of encoding a continuous signal into spikes, \ac{BSA} assumes the signal has already been transformed into a spike train. The objective is to decode this spike train with minimal reconstruction error. The encoding process is assumed to involve convolution with a \ac{FIR} filter, and decoding requires reversing this operation through deconvolution. At each step, an error term is computed to measure the sum of the differences between the filter and the signal, while a second error term quantifies the sum of the signal itself. Spikes are emitted when the first error term is smaller than the second minus a threshold value. Upon spike emission, the filter is subtracted from the signal. A pseudocode implementation of \ac{BSA} encoding is provided in Figure \ref{fig:BSA_encoding} (based on \cite{BSA_SOURCE_schrauwen2003bsa}).


\begin{figure}[htbp] 
    \centering 
    \begin{minipage}{1\linewidth} 
        \footnotesize 
        \begin{algorithmic}[1] 
            \State \textbf{Input:} $signal$, $filter\_order$, $filter\_cuto\textit{ff}$, $threshold$
            \State \textbf{Output:} $spike\_train$
            \State $signal \gets \text{normalize}(signal)$
            \State $fir\_coe\textit{ff} \gets \text{fir\_filter}(filter\_size = filter\_order + 1, filter\_cuto\textit{ff}, sampling\_frequency = 1)$
            \State $spikes \gets 0$
            \For{$t = 1$ to $time\_steps$}
                \State $err1 \gets 0$
                \State $err2 \gets 0$
                \For{$j = 1$ to $filter\_size$}
                    \If{$t + j - 1 \leq time\_steps$}
                        \State $err1 \gets err1 + |signal(t + j - 1) - fir\_coe\textit{ff}(j)|$
                        \State $err2 \gets err2 + |signal(t + j - 1)|$
                    \EndIf
                \EndFor
                \If{$error1 \leq err2 - threshold$}
                    \State $spikes(t) \gets 1$
                    \For{$j = 1$ to $filter\_size$}
                        \If{$t + j - 1 \leq time\_steps$}
                            \State $signal(t + j - 1) \gets signal(t + j - 1) - fir\_coe\textit{ff}(j)$
                        \EndIf
                    \EndFor
                \Else
                    \State $spikes(t) \gets 0$
                \EndIf
            \EndFor
        \end{algorithmic}
    \end{minipage} 
    \caption{BSA Encoding Method} 
    \label{fig:BSA_encoding} 
\end{figure}

Finally, \ac{PWM} is based on comparing the input signal to a carrier (or "reference") signal and emitting spikes whenever the carrier signal exceeds the input signal. Once this condition is met, the input signal must first fall below the carrier signal before exceeding it again for a new spike to be emitted. The carrier signal is a sawtooth wave with an adjustable frequency parameter. As with the other encoding methods, the input signal must be normalized to ensure it overlaps with the carrier signal. A pseudocode implementation of the \ac{PWM} encoding method is shown in Figure \ref{fig:PWM_encoding} (based on \cite{PWM_SOURCE_arriandiaga2019pulsewidth}).\\


\begin{figure}[htbp] 
    \centering 
    \begin{minipage}{1\linewidth} 
        \footnotesize 
        \begin{algorithmic}[1] 
            \State \textbf{Input:} $signal$, $frequency$, $downspike$ (boolean)
            \State \textbf{Output:} $spike\_train$
            \State $signal \gets \text{normalize}(signal)$
            \State $carrier \gets \text{sawtooth}(frequency)$
            \State $neg\_carrier \gets 1 - carrier$
            \State $pwm \gets 0$, $up\_spikes \gets 0$, $down\_spikes \gets 0$
            \For{$t = 1$ to $time\_steps$}
                \If{$signal(t) < carrier(t)$}
                    \State $pwm(t) \gets 1$
                \ElsIf{$signal(t) > neg\_carrier(t)$ \textbf{and} $downspike = \text{True}$}
                    \State $pwm(t) \gets -1$
                \Else
                    \State $pwm(t) \gets 0$
                \EndIf
            \EndFor
            \For{$t = 2$ to $time\_steps$}
                \If{$pwm(t) = 1$ \textbf{and} $pwm(t-1) \neq 1$}
                    \State $up\_spikes(t) \gets 1$
                \ElsIf{$pwm(t) = -1$ \textbf{and} $pwm(t-1) \neq -1$}
                    \State $down\_spikes(t) \gets -1$
                \EndIf
            \EndFor
            \State $spikes \gets up\_spikes + down\_spikes$
        \end{algorithmic}
    \end{minipage} 
    \caption{PWM Encoding Method} 
    \label{fig:PWM_encoding} 
\end{figure}

\subsection{Experimental setup}
To evaluate the performance characteristics, such as runtime and power consumption, we utilized specific embedded hardware. While the core encoding algorithms are implemented in C++, the subsequent performance results presented are specific to the chosen platform and configuration. We use an MCX-N947-EVK development board from NXP Semiconductors as an embedded evaluation platform. 
The MCX N is a multi-core System on Chip comprising two Arm Cortex®-M33 cores, 2MB flash, 512kB SRAM, a digital signal co-processor, and a neural processing unit for accelerating inference of conventional neural networks.
One Cortex-M33 has a maximum clock speed of 150MHz and offers a favorable trade-off between energy efficiency, real-time determinism, and system security. 
The power consumption comes down to 57 µA/MHz in active mode. 
For our evaluation, we configured the system’s clock frequency to 12 MHz and set the power mode to overdrive with a DCDC core voltage of 1.2V, balancing power consumption and performance.

The interaction with the development board is facilitated by eRPC calls, wherein the board is treated as a server and the PC as a client. First, the client transmits its signal data to the board via UART. It then requests the execution of the selected encoding function on the transmitted data. Finally, the client requests the results once encoding is completed, and the board transfers them back. All measurements are performed between the start and end of the encoding method to exclude data transfer and communication overhead. Additionally, the normalization operations required by some methods have been excluded.

The power consumption of each method was measured over 10 hours at a sample rate of 62500 samples per second, with the results averaged over the entire duration. It should be noted that even when no operation is performed, the board continues to consume power. Therefore, a baseline power consumption measurement was first taken and subsequently subtracted from each method's measurements, allowing for a direct comparison between the encoding methods. The time measurements were taken using C functions, which initiated and terminated a timer at the beginning and end of method execution, respectively. 

In order to quantify the extent of information loss incurred during the encoding process, we calculate the reconstruction error for each encoding method by decoding the encoded signal and measuring the \ac{MSE} between the reconstruction and the original signal.

\subsection{Evaluated Signals}

This study selected four artificially generated signals with 16384 time steps, each normalized to unit mean and variance, to assess the encoding methods. These specific signals were chosen to represent a diverse set of common temporal dynamics, including irregularity, drift, abrupt transitions, and smooth periodicity.

\begin{figure}[t]
\centering
\begin{tikzpicture}
\node[rotate=90, anchor=south] at (-1, -0.0875\textwidth) {Amplitude};

\begin{groupplot}[
    group style={
        group size=1 by 4,
        horizontal sep=0cm,
        vertical sep=0.2cm,
        xticklabels at=edge bottom,
        xlabels at=edge bottom,
    },
    width=0.45\textwidth,
    height=0.175\textwidth,
    xlabel={Step},
    grid=major
]

\nextgroupplot
\addplot[thin, blue] table[x=step,y=amplitude,col sep=comma]{misc/data/signal_vibration.csv};
\node at (rel axis cs:0.1,0.9) {\textbf{(a)}};

\nextgroupplot
\addplot[thin, blue] table[x=step,y=amplitude,col sep=comma]{misc/data/signal_trended.csv};
\node at (rel axis cs:0.1,0.9) {\textbf{(b)}};

\nextgroupplot
\addplot[thin, blue] table[x=step,y=amplitude,col sep=comma]{misc/data/signal_rectangular.csv};
\node at (rel axis cs:0.1,0.9) {\textbf{(c)}};

\nextgroupplot
\addplot[thin, blue] table[x=step,y=amplitude,col sep=comma]{misc/data/signal_sinusoidal.csv};
\node at (rel axis cs:0.1,0.9) {\textbf{(d)}};

\end{groupplot}
\end{tikzpicture}
\caption{Four different signal types: (a) Vibration Signal, (b) Trended Signal, (c) Rectangular Signal, (d) Sinusoidal Signal.}
\label{fig:sig}
\end{figure}
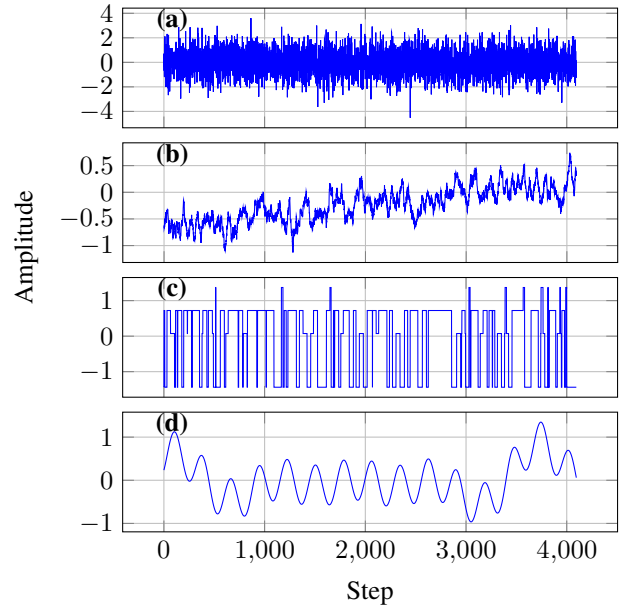

The "Vibration" signal (Figure \ref{fig:sig} (a)) simulates oscillations with a specific standard deviation and frequent fluctuations, challenging the encoders due to its irregularity. The "Trended" signal (Figure \ref{fig:sig} (b)) introduces drift, testing the stability of the reconstruction over time. The "Rectangular" signal (Figure \ref{fig:sig} (c)) resembles a square wave, providing insights into the methods' performance on constant and abruptly transitioning values. Lastly, the "Sinusoidal" signal (Figure \ref{fig:sig} (d)), smooth and noise-free, assesses the preservation of periodic patterns.

\subsection{Paramter Optimization}
Each encoding method was optimized through 500 trials. In each trial, the chosen set of parameters was used to encode the signal into spikes. The resulting spike train was then decoded to reconstruct the original signal, and the reconstruction error, quantified as the \ac{MSE} between the original and reconstructed signals, was computed. The optimization aimed to minimize this reconstruction error by adjusting the encoding parameters.

To perform this optimization, we employed Optuna \cite{IMP_optuna}, a widely used hyperparameter optimization framework implemented in Python. For most encoding methods, random sampling was used to explore the parameter space randomly. However, for the BSA method, which involves tuning three interdependent parameters and is computationally demanding, the TPESampler \cite{ozaki2020multiobjective} was utilized due to its efficiency in complex, multi-variable optimizations.

\section{Results}
    \label{sec:results}
This section presents the results for each encoding method applied to different signal types, evaluated using four main criteria: reconstruction error (TABLE \ref{tab:reconstruction_error}), encoding time, power consumption (TABLE \ref{tab:time_power_measurement}), and spike sparsity (TABLE \ref{tab:numspikes_to_timesteps_ratio}). Spike sparsity is expressed as the ratio of spikes to the total signal length ($16384$).\

The absolute power values reported in TABLE \ref{tab:time_power_measurement} reflect the average power consumption measured over ten hours for the entire board. Additionally, the absolute power was measured for the no-operation (NOP), resulting in a value of \SI{99.74}{\nano\watt}. The NOP reference power is subtracted from this value to determine the dynamic power consumed by each encoding method. The last three rows of the table indicate the encoding time, dynamic power, and dynamic energy consumption for each method, expressed as a percentage increase relative to the SF converter, which shows the lowest values across all methods.


\begin{table}[t!] \centering \caption{Reconstruction Error (MSE)} 
\label{tab:reconstruction_error} 
\centering \begin{tabular}{|l|c|c|c|c|} 
\hline 
Encoding Method & LIF & SFC & PWM & BSA \\
\hline
Vibration & \textbf{0.370641} & 0.487395 & 1.099153 & 0.845590 \\
Trended & 0.102157 & \textbf{0.000970} & 0.015004 & 0.006510 \\
Rectangular & 0.081864 & 0.157202 & 0.172042 & \textbf{0.063650} \\
Sinusoidal & 0.126749 & \textbf{0.000139} & 0.004631 & 0.013472 \\

\hline Mean Error & 0.170353 & \textbf{0.161427} & 0.322707 & 0.232305 \\
\hline 
\end{tabular} 
\end{table}


\begin{table}[t!] 
\centering 
\caption{Spike Sparsity} 
\label{tab:numspikes_to_timesteps_ratio} 
\centering 
\begin{tabular}{|l|c|c|c|c|} 
\hline 
Encoding Method & LIF (\%) & SFC (\%) & PWM (\%) & BSA (\%) \\ 
\hline 
Vibration & \textbf{36.83} & 41.26 & 50.00 & 53.81 \\
Trended & 65.59 & 35.82 & \textbf{3.66} & 61.34 \\ 
Rectangular & 62.47 & \textbf{10.20} & 14.33 & 48.48 \\ 
Sinusoidal & 42.72 & 35.14 & \textbf{3.02} & 48.70 \\ 

\hline 
Mean Sparsity & 51.90 & 30.60 & \textbf{17.75} & 53.08 \\ 
\hline 
\end{tabular} 
\end{table}

\begin{table}[t!]
    \centering
    \caption{Time, Power and Energy Measurements}
    \label{tab:time_power_measurement}
\centering
\begin{tabular}{|l|c|c|c|c|}
\hline
Encoding Method & LIF & SF & PWM & BSA \\
\hline
time (\SI{}{\milli\second}) & 127.15 & \textbf{123.52} & 691.32 & 2238.58 \\
absolute power (\SI{}{\nano\watt}) & 109.45 & \textbf{106.61} & 111.15 & 119.74 \\
dynamic power (\SI{}{\nano\watt}) & 9.71 & \textbf{6.87} & 11.41 & 20.00 \\
dynamic energy (\SI{}{\pico\watt\hour}) & 0.34 & \textbf{0.24} & 2.19 & 12.44 \\
\hline

time wrt. SF (\%) & 2.94 & \textbf{0.00} & 459.64 & 1712.28 \\
dynamic power wrt. SF (\%) & 41.34 & \textbf{0.00} & 66.08 & 191.12 \\
dynamic energy wrt. SF (\%) & 41.67 & \textbf{0.00} & 812.50 & 5083.33 \\
\hline
\end{tabular}
\end{table}

\section{Discussion}
    \label{sec:discussion}
This study evaluated four encoding methods—LIF, SF, PWM, and BSA—based on their performance in reconstructing signals, spike sparsity, and energy efficiency.

LIF demonstrated high accuracy with vibration data due to its suitability for signals fluctuating around a baseline. However, its mean bias hampered performance with trended signals and capturing extreme values in sinusoidal signals. Despite a high spike rate, its simple implementation resulted in good energy efficiency, aligning with findings from Yarga et al. \cite{yarga2022efficient} on its suitability for fluctuating signals.

The SF Converter generally achieved the lowest reconstruction error, especially with sinusoidal signals, and offered superior speed and energy efficiency. However, it exhibited instability with rectangular waveforms. Our results confirm prior evidence from Chen et al. \cite{chen2022temporal} on SF’s accuracy, showing it had the lowest mean reconstruction error (0.1614 - TABLE \ref{tab:reconstruction_error}) and minimal power consumption (TABLE \ref{tab:time_power_measurement}), underscoring its efficiency and versatility.
 
PWM, despite low spike sparsity, underperformed in reconstruction accuracy, struggling with fluctuating and rectangular waveforms. Its complex operations resulted in higher encoding time and power consumption.  This contrasts with Wang et al. \cite{wang2023comparison}, who reported favorable reconstruction and power consumption for PWM in simpler implementations. Our results indicate PWM’s higher reconstruction error (mean MSE of 0.3227) and elevated power consumption (812.5\% relative to SF's energy usage), likely due to its sensitivity to parameter tuning.

BSA effectively filtered high frequencies and achieved low reconstruction error for rectangular signals (MSE of 0.0636) but suffered from high computational costs and sensitivity to initial signal values, consistent with Wang et al. \cite{wang2023comparison}. This limits its practicality for real-time embedded systems, as its encoding duration and power consumption are significantly higher than SF.

Overall, SF emerged as the most energy-efficient and reliable method, while LIF and BSA provide niche strengths in specific applications. Future work may benefit from investigating hybrid approaches.

\section{Spike Encoding Framework}
   \label{sec:repository}
Our framework is based on the idea of a Converter. In our approach, a Converter is an object that can encode and decode signals. Optionally, Converters may also include a method of optimization that allows them to finetune their hyperparameters to a specific signal. In this way, each encoding method can achieve its highest reconstruction accuracy without the need to finetune its parameters manually.

For example, consider using our \ac{SF} implementation depicted in \autoref{fig:sf_encoding_list}. \ac{LIF}, \ac{PWM} and \ac{BSA} encoding all function analogously. 

\begin{figure}[t!]
\centering
    \begin{minipage}{0.45\textwidth}
        \begin{lstlisting}
import torch
from encoding.step_forward_converter import StepForwardConverter

signal = torch.tensor([0.1, 0.3, 0.2, 0.4, 0.8])

converter = StepForwardConverter()
spikes = converter.encode(signal)
        \end{lstlisting}
    \end{minipage}
    \caption{Using the StepForwardConverter to encode a signal.}
    \label{fig:sf_encoding_list}
\end{figure}

Decoding spike trains is achieved similarly, except that \textit{encode} is replaced with \textit{decode}. In \autoref{fig:sf_decoding_list}, we see how converters can be optimized and used to decode signals.

\begin{figure}[t!]
\centering
\begin{minipage}{0.45\textwidth}
    \begin{lstlisting}
# ... same as above ...
optimalthreshold = converter.optimize(signal)
optimized_converter = StepForwardConverter(optimalthreshold)

spikes = optimized_converter.encode(signal)
reconstructed_signal = optimized_converter.decode(spikes)
    \end{lstlisting}
    \end{minipage}
    \caption{Minimal example of optimizing a converter and decoding signals in order to reconstruct them.}
    \label{fig:sf_decoding_list}
\end{figure}

Additionally, we introduce a custom encoder designed explicitly for Gymnasium \cite{gym_towers2024gymnasium} environments, the GymnasiumEncoder. It extends rate encoding methods by adding utility methods tailored for reinforcement learning tasks. We also provide the BinEncoder, which utilizes \ac{GRF} encoding to transform single values into multiple bin responses. However, like the GymnasiumEncoder, it is restricted to encoding operations only.

\section{Conclusion and Future Work}
    This paper introduces a novel, open-source PyTorch-compatible Python framework for spike encoding, explicitly designed for machine learning and reinforcement learning applications. The framework offers support for a diverse range of encoding methods, encompassing conventional algorithms, such as LIF, SF, PWM, and BSA, as well as specialized components like a reinforcement learning-optimized encoder and Gaussian Receptive Field-based population coding. The framework is accompanied by documentation and testing, ensuring seamless integration into machine learning workflows and fostering accessibility and ease of use.

Furthermore, a comprehensive evaluation of the performance trade-offs of each encoding method was conducted by implementing them in C++ and testing them on embedded hardware. Our findings indicate that while SF exhibited the highest energy efficiency and fastest encoding time, it tended to falter in abrupt signal transitions. LIF demonstrated efficacy in handling fluctuating signals but exhibited limitations in the presence of trends or extreme values. PWM demonstrated lower accuracy and higher energy consumption than the other methods. In contrast, BSA demonstrated high accuracy for certain signal types and filtering capabilities but at the cost of increased computational demands. These comparative insights provide valuable guidance for selecting the most suitable encoding method, supporting the broader adoption of Spiking Neural Networks in machine learning applications.

Future work will focus on extending the framework's capabilities by implementing additional encoding algorithms, including both temporal and rate coding approaches, to allow for broader comparisons.

\section*{ACKNOWLEDGMENTS} This research is funded by the German Federal Ministry of Education and Research as part of the project ”ThinKIsense“, funding no. 16ME0564.
    
\section*{Supplementary Materials}
We provide Python implementations for all the investigated algorithms. Our repository includes \ac{LIF} encoding, \ac{SF}, \ac{PWM}, and \ac{BSA}, as well as two additional algorithms. The first is a custom encoder particularly suited to reinforcement learning, especially to Gymnasium \cite{gym_towers2024gymnasium} environments. The second one implements a form of population coding that is based on Gaussian Receptive Fields. Our code can be accessed at \cite{vasilache_alex-vasilachespike-encoding_2025}.

\printbibliography

\end{document}